\documentclass[runningheads]{llncs}
\usepackage{graphicx}
\usepackage{algorithm}
\usepackage{algpseudocode}
\usepackage{enumitem}
\usepackage{caption}
\usepackage{subcaption}
\usepackage{wrapfig}
\usepackage{color,soul}
\newlist{level}{itemize}{4}
\setlist[level]{label={},noitemsep,topsep=0pt}
\usepackage{amsmath}
\DeclareMathOperator*{\argmax}{argmax}
\usepackage{amssymb}
\usepackage{amstext}

\begin{document}
%
\title{Staged Reinforcement Learning for Complex Tasks through Decomposed Environments}

%
\titlerunning{Staged RL for Complex Tasks through Decomposed Environments}
%
\author{Rafael Pina\inst{1}\orcidID{0000-0003-1304-3539} \and
Corentin Artaud\inst{1}\orcidID{0009-0002-0387-235X} \and
Xiaolan Liu\inst{1}\orcidID{0000-0002-7500-9128} \and
Varuna De Silva\inst{1}\orcidID{0000-0001-7535-141X}}
\authorrunning{R. Pina et al.}
%
\institute{$^1$Institute for Digital Technologies, Loughborough University London, 3 Lesney Avenue E20 3BS, London, United Kingdom
\email{\{r.m.pina,c.artaud2,xiaolan.liu,v.d.de-silva\}@lboro.ac.uk}}
\maketitle              
\begin{abstract}
Reinforcement Learning (RL) is an area of growing interest in the field of artificial intelligence due to its many notable applications in diverse fields. Particularly within the context of intelligent vehicle control, RL has made impressive progress. However, currently it is still in simulated controlled environments where RL can achieve its full super-human potential. Although how to apply simulation experience in real scenarios has been studied, how to approximate simulated problems to the real dynamic problems is still a challenge. In this paper, we discuss two methods that approximate RL problems to real problems. In the context of traffic junction simulations, we demonstrate that, if we can decompose a complex task into multiple sub-tasks, solving these tasks first can be advantageous to help minimising possible occurrences of catastrophic events in the complex task. From a multi-agent perspective, we introduce a training structuring mechanism that exploits the use of experience learned under the popular paradigm called Centralised Training Decentralised Execution (CTDE). This experience can then be leveraged in fully decentralised settings that are conceptually closer to real settings, where agents often do not have access to a central oracle and must be treated as isolated independent units. The results show that the proposed approaches improve agents performance in complex tasks related to traffic junctions, minimising potential safety-critical problems that might happen in these scenarios. Although still in simulation, the investigated situations are conceptually closer to real scenarios and thus, with these results, we intend to motivate further research in the subject.

\end{abstract}
\section{Introduction}\label{sec:intro}
Reinforcement Learning (RL) is a popular subject in the field of Machine Learning whose many notable applications have raised interest within the scientific community. In fields ranging from robotics, gaming or healthcare, to autonomous vehicles or finance, RL has been proved to have relevant applications \cite{hester_robotics_2012,hu_finrl_2019,pineau_epilepsy_2009}. 

Due to the constraints that can be lifted in the world of computer games, the success of RL in this area is particularly notorious. For example, works such as \cite{silver_2016_mastering} describe how RL agents have defeated the human being playing games like Go or achieved super-human performance in multiple Atari Games \cite{mnih_2015_humanlevel}. However, when applying the same concepts to real-world scenarios, studies have demonstrated how challenging it can be to reproduce the super-human behaviours learnt in simulation \cite{dulacarnold_2019_challenges}. Naturally, the real-world can be very complex and it is often unfeasible to encounter in simulation all the possibilities that may appear in real life. Recent works showed how this can be achieved in controlled scenarios, where the reality is very close to the simulation, making the tasks easier \cite{kalapos_2020,tobin_2017,almasi_2020}. However, this procedure is not always feasible outside controlled environments, since it is impossible to predict with maximum accuracy what will happen in our lives as human beings living in an extremely dynamic and stochastic world.

In the context of autonomous vehicle control, there is a wide research that focuses on approximating the use of RL in the real-world \cite{almasi_2020,pina_2021,kober_2013,anirudth_2020}. One of the solutions to facilitate learning complex situations is to break complex tasks into simpler tasks. If a given agent starts learning from simpler steps, then it is possible to mix this intermediate knowledge to address a more complex task that can be built from the simpler parts learned. Additionally, when learning simpler parts of a complex task, agents are less likely to cause safety problems when it comes to real applications, since they do not need to attempt and probably fail many times a very complex task that would require a long learning time.

\begin{figure}[!t]
    \centering
    \includegraphics[width=0.9\textwidth]{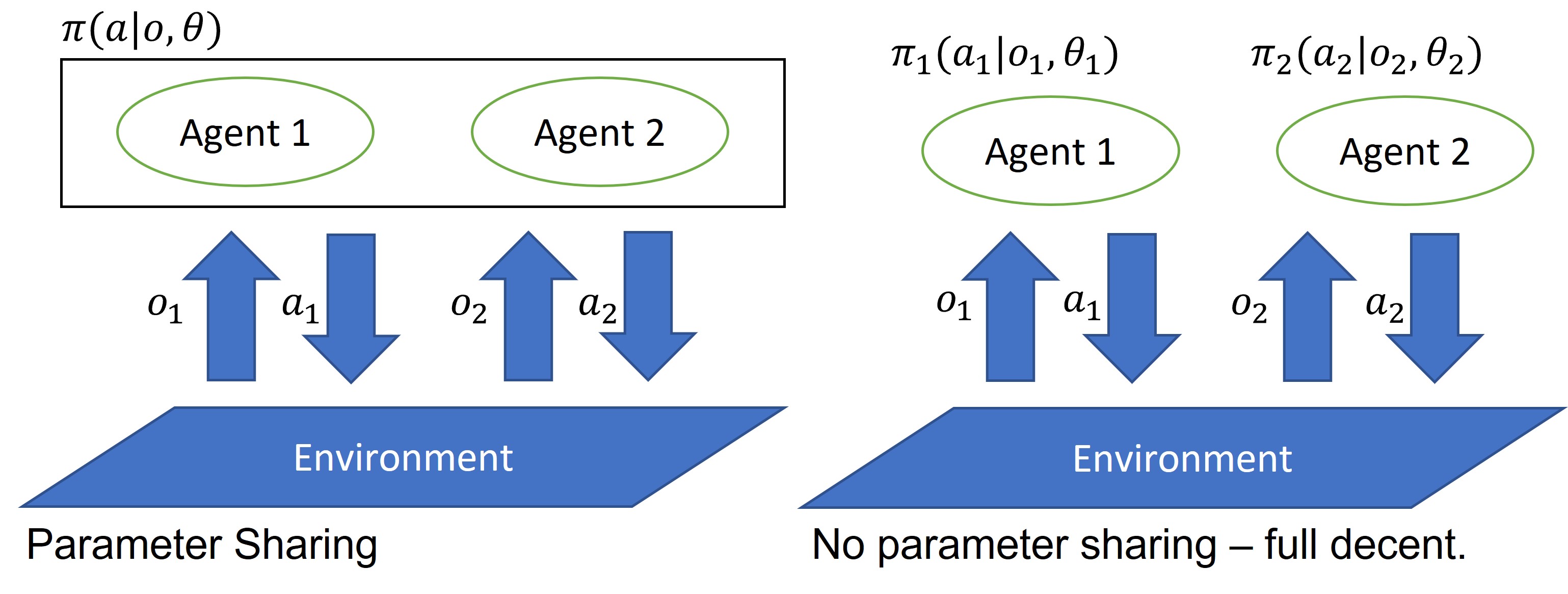}
    \caption{Illustration of the main differences between sharing parameters (left) and not sharing parameters (right) of the agent networks in MARL.}
    \label{fig:param_share_vs_no_param_share}
\end{figure}

While these problems have been deeply studied in single-agent settings, it is important to note how multi-agent systems are gaining attention, and RL has been proved to be a suitable solution to address challenges that require multiple agents to interact \cite{nguyen_2020}. Multi-Agent Reinforcement Learning (MARL) is the sub-field of RL that studies its applications in muti-agent systems. Learning how to interact as a group brings endless challenges worth investigating. Multiple recent works have shown the potential of using MARL, mostly in controlled simulated environments. Famous games such as StarCraft II \cite{starcraft_2019} or the Google Football environment \cite{kurach_2020_google} have served as the workstations for training and studying MARL agents. Other works such as \cite{wang_qplex_2021,ruan_2022} demonstrated how MARL can learn complex policies in these kind of computer simulated games. A large part of approaches proposed to tackle these problems follow a general paradigm named Centralised Training Decentralised Execution (CTDE) \cite{olienoek_2008,KRAEMER201682}. With this configuration, the agents have access to extra information of the environment during training, but are constrained to their local observations when executing their actions. Yet, when looking at real-world applications, it is often impossible to provide a centralised oracle to the agents that allows them to see the full state of the environment at any moment during their training phases. Another popular convention in these methods that tackle problems in simulated MARL is to share the network parameters of the learning agents (Fig. \ref{fig:param_share_vs_no_param_share} illustrates the differences between sharing and not sharing parameters in MARL). Logically, sharing a network through a big team of agents in reality can bring endless problems. Communication-based approaches can be seen as a solution for MARL in real applications, but in safety critical situations, even very small latency periods can cause tragic events that must be prevented \cite{sun_latency_2020}.  

In this paper, we investigate different applications of RL and MARL in simulated environments in the context of vehicle control, while discussing how we can create methods that are closer to real-world applications. We study the problems of task decomposition in the context of obstacle avoidance and goal reaching, and also demonstrate a method for mixing simpler tasks into the respective complex task. If agents can learn efficiently in a staged manner, learning in reality can also be easier by following simpler steps in order to achieve a more difficult goal, and avoiding potential collisions that could happen if they spend too long learning a difficult task straight away. In addition, we step into the world of multi-agent systems and discuss a training framework in MARL that consists of training agents in simulation following the advantageous CTDE paradigm, and then reuse the learnt policies in a more complex environment that is fully decentralised. By structuring the learning process following this procedure, agents can take advantage of being trained under the CTDE paradigm, and then can be used in more complex settings, and retrained in a fully decentralised basis. The results suggest that this training framework accelerates learning in more complex simulated environments, reducing potential safety problems that would incur from collisions between agents. In this sense, the contributions in this paper are as follows:
\begin{itemize}
    \item We show that certain simpler tasks can be mixed into more complex tasks to improve learning in the context of obstacle avoidance and goal reaching.
    \item We investigate a training framework that firstly consists of training agents under the advantageous CTDE paradigm with parameter sharing. Then, the learnt policies can be transferred to more complex environments and continue to be trained but in a fully decentralised setting (no parameter sharing) that is more suitable for future real applications.
\end{itemize}

\section{Related Work}\label{sec:rel_work}
There are several practical problems in adopting RL in real applications, especially in safety critical situations. RL typically requires many iterations of trial and error to learn a given task. Although such trial and error is performed in a simulated virtual environment, transferring the intelligent capabilities learnt in simulation into the physical real world poses a significant challenge \cite{kalapos_2020}. Another problem is that mistakes by robots in physical settings could lead to property damages and even costly damages to human lives \cite{almasi_2020,dulacarnold_2019_challenges}. Therefore, safety critical guarantees of such RL based controllers are required to ensure that RL agents can perform safely.  

It is clear how training agents in a controlled simulated environment is a key factor to create safe entities. For instance, in the concept of Multi-Fidelity Reinforcement Learning (MFRL) \cite{suryan_2020}, an RL agent is trained in multiple simulators of the real environment at varying fidelity levels. It is demonstrated that by increasing simulation fidelity, the number of samples used in successively higher simulators can be reduced \cite{chebotar_2019}.

In \cite{kalapos_2020}, the authors demonstrated that a vision-based lane following and obstacle avoidance RL agent can learn a suitable steering and obstacle avoidance policy in simulation that then can be transferred to a simple physical setting. However, it is important to note how the used scenarios are controlled and forced to minimise the reality gap (difference between simulation and reality). In \cite{kalapos_2020}, the authors concluded that some trained RL agents can be sensitive to problem formulation aspects such as the representation of real world actions. It was shown that, by using domain randomization, a moderately detailed and accurate simulation is sufficient for training such an agent that can operate in a real environment. Domain randomization is another method that minimises the reality gap, allowing to make RL policies learned in simulation closer to reality \cite{tobin_2017}.

With the proven advantages and success of using simulated environments to analyse safety critical problems, the literature regarding approaches that tackle RL problems in more complex simulated environments encompasses a broad range \cite{wang_qplex_2021,mnih_2015_humanlevel,kurach_2020_google,sunehag_value-decomposition_2018}. Works such as \cite{lerer_convention_2019} analyse how RL agents learn how to act in traffic environments in a safe way. In particular, the authors show how agents learn conventions that are compatible with human rules of driving. In the more complex multi-agent setting, MARL has been well studied. In the work of \cite{chu2020multiagent}, the authors show how different agents learn to navigate in traffic scenarios and learn driving policies in traffic networks. Under multi-agent settings, the CTDE paradigm is important to scale the approaches \cite{qtran_2019}. Several works have demonstrated how this setting brings benefits to the performances of trained agents in multi-agent settings that must cooperate \cite{sunehag_value-decomposition_2018,qtran_2019,wang_qplex_2021}. In addition, sharing the parameters of multiple agents in multi-agent scenarios is also a convention widely used in the literature \cite{gupta_2017}. Although following such procedures can be very profitable in simulation, it is very challenging to establish a reliable CTDE configuration in real applications, or to have a central entity that allows to share the same network across a number of agents. Instead, the fully decentralised setting sounds much more inviting, since each agent can be treated as an isolated independent entity.
\begin{algorithm}[!t]
\caption{Algorithm 1}
Let $T$ be a certain decomposable task
\\
Decompose $T$ in $n$ sub-tasks, $\{T_1,\dots,T_n\}$
\\
\textbf{for each} sub-task in $\{T_1,\dots,T_n\}$
\begin{level}
    \item Train $Q$-table $Qt_i$
    \item Save $Q$-table $Qt_i$
\end{level}
\textbf{end for}
\\
Initialise empty new table $Qt_{jt}$
\\
\textbf{for each} $Q$-table in $\{Qt_1,\dots,Qt_n\}$
\begin{level}
    \item \textbf{if} $Qt_{jt}$ is empty \textbf{then} 
    \begin{level}
        \item \textbf{Add} all $Qt_i$ entries
    \end{level}
    \textbf{else if} $Qt_i$ entries in $Qt_{jt}$ \textbf{then}
    \begin{level}
        \item \textbf{Combine} entries
    \end{level}
    \textbf{else}
    \begin{level}
        \item \textbf{Pass}
    \end{level}
\end{level}
\textbf{end for}
\\
\textbf{Output} joint $Q$-table $Qt_{jt}$
\label{alg:alg_1}
\end{algorithm}

\section{Methods}\label{sec:meth}
\subsection{Task Factorization}\label{sec:meth_task_fact}
In this section, we introduce a method for solving single-agent tasks using RL, based on the assumption that a certain complex task can be factorized into multiple simple sub-tasks. These sub-tasks are seen as intermediate objectives that must be learned to solve a global task. In this sense, we train multiple Q-tables in separate to solve sub-tasks that together form a main complex task. By learning these tables, it is possible to create a joint Q-table that contains a mix of the Q-tables corresponding to the learned sub-tasks. The agents use Q-learning \cite{watkins_1992_technical} to learn the Q-tables. This algorithm is part of the foundations of RL, where the updates to optimise a Q-function to solve a certain problem are made by following the update rule
\begin{equation}\label{eq:q_up}
    Q(s,a)=(1-\alpha)Q(s,a)+\alpha\left[r+\gamma\mathop{\mathrm{max}}_{a'}Q(s',a')\right]
\end{equation}
for a certain state and action pair $(s, a)$, and the corresponding next pair $(s',a')$, where $\alpha$ is a learning rate, $\gamma$ a discount factor, and $r$ the reward received.

In summary, this task decomposition procedure can be formalized as the following: given a certain task $T$, it is possible to decompose it into $n$ sub-tasks. Thus, $n$ Q-tables can be learned separately and then exported to form a joint Q-table, $Qt_{jt}$, which is a combination of all the sub Q-tables (as described in Algorithm \ref{alg:alg_1}).

To demonstrate our assumption, we show ahead in the results section how a complex task can be solved using the joint Q-table that arises from this procedure. This table results from the combination of the separate Q-tables trained in the sub-tasks in which the complex task was decomposed. Hence, the joint Q-table will perform in an environment where it was not trained in advance.
\begin{wrapfigure}[20]{rt}{0.5\textwidth}
    \centering
    \includegraphics[width=0.5\textwidth]{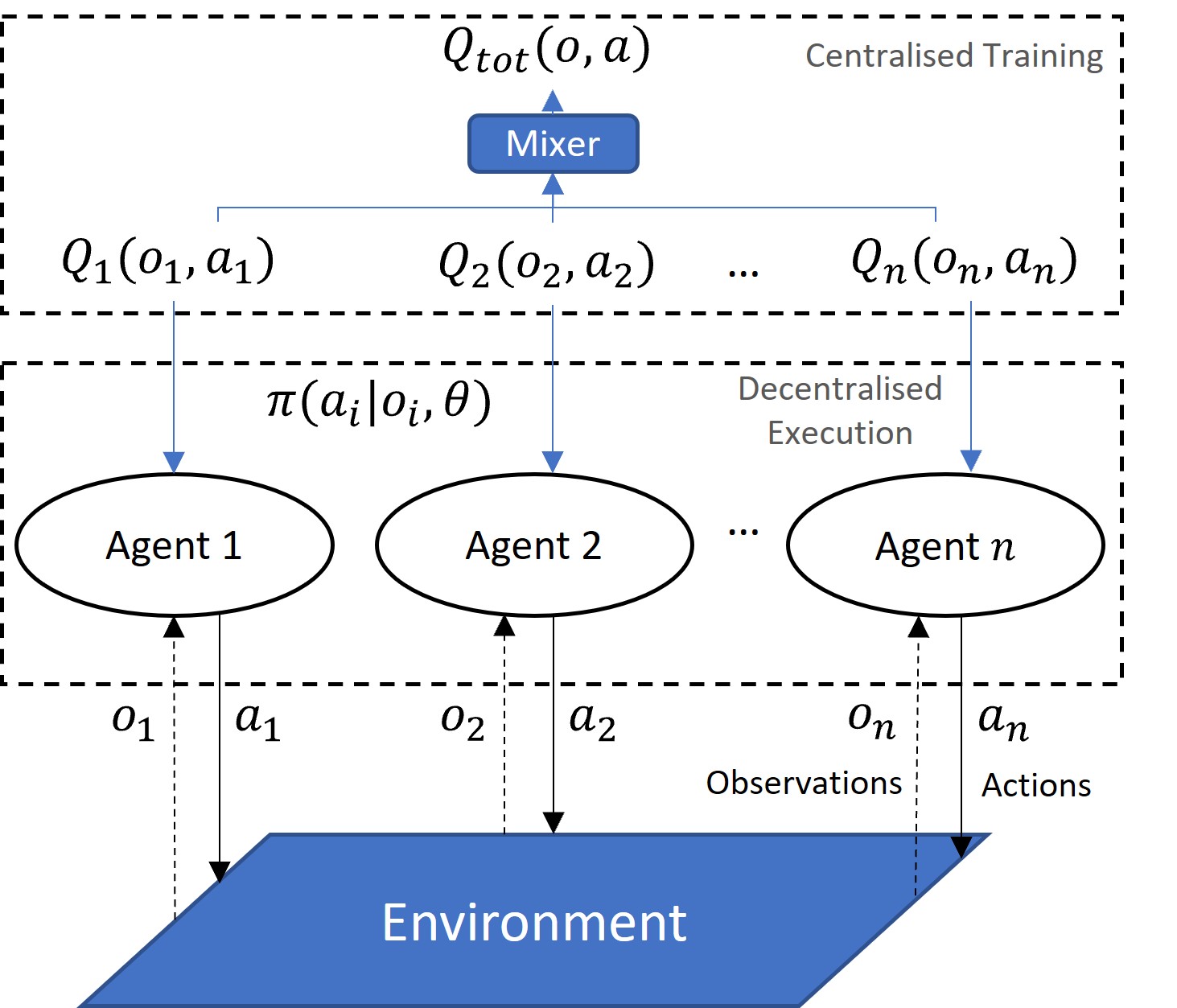}
    \caption{A simplified overview of how value function factorization methods operate under the CTDE paradigm, considering pairs of observations and actions.}
    \label{fig:ctde_overview}
\end{wrapfigure}

\subsection{From CTDE to Full Decentralisation}\label{sec:meth_ctde_decent}
The second approach that we discuss in this paper is related to MARL problems. This approach consists of a structured training procedure that is composed by two main stages that will be described in detail in this section. In the first stage, we train a team of agents in a simpler environment under the popular CTDE paradigm with parameter sharing. To do so, we use Value Decomposition Networks (VDN) \cite{sunehag_value-decomposition_2018}, a popular algorithm that is part of a big family of MARL value function factorization algorithms. The key idea of this family of methods is to learn a factorization of a joint Q-function $Q_{tot}$ into a set of individual Q-functions corresponding to each one of the agents in the multi-agent team. In the case of VDN, this factorization is achieved through addition and can be formally represented by the equation
\begin{equation}\label{eq:3}
    Q_{tot}\left(\tau ,a\right)\mathrm{=}\sum^N_{i\mathrm{=1}}{Q_i\mathrm{(}\tau_i,a_i\mathrm{;}{\theta }_i\mathrm{)}}
\end{equation}
for a set of action-observation histories and action pairs $(\tau_i,a_i)$, and the $\theta$ values are the parameters of the agent neural networks. 

To learn this factorization, these methods use this mixer that mixes individual Q-values into the joint Q-value. This mixer can be very different from method to method and it is the key factor to the type of Q-functions that each method can factorise (representational complexity). Formally, a value function factorization method is said to be effective if it satisfies the Individual-Global-Max (IGM) \cite{qtran_2019} condition
\begin{equation}\label{eq:2}
\argmax_aQ_{tot}\left(\tau,a\right)=(\argmax_{a_1}Q_1(\tau_1,a_1),\dots,\argmax_{a_N}Q_N(\tau_N,a_N))
\end{equation}
In simple terms, this condition means that, for an efficient factorisation, the set of local optimal actions should maximise the joint $Q_{tot}$. Fig. \ref{fig:ctde_overview} depicts an overview of how value function factorisation methods operate with parameter sharing and under the CTDE paradigm. 

The second stage starts after training the agents in the simpler environment following the procedure described above. In this stage, we transfer the trained agents to a more complex environment that is now fully decentralised. This means that, on top of the environment being more complex, the agents only have access to their local observations and do not share parameters anymore (as summarised in Fig. \ref{fig:param_share_vs_no_param_share}). In the considered full decentralised setting, each agent is treated as an independent entity. Thus, we use the famous DQN method \cite{mnih_2015_humanlevel} to control each one of the agents as an independent agent. Since we do not share parameters, beyond not using any CTDE method in this stage, each agent has its own DQN controller. We refer to this method ahead as IDQL (Independent Deep Q-learning). Following the authors in \cite{mnih_2015_humanlevel}, agents using this method learn by minimising the loss  
\begin{equation}
    \mathcal{L}(\theta)=\mathbb{E}_{b\sim B}\left[\big(r+\gamma\mathop{\mathrm{max}}_{a'}Q(\tau',a';\theta^-)
    -Q(\tau,a;\theta)\big)^2\right]
\end{equation}
where $b$ represents a set of experience samples from a replay buffer $B$ and $\theta^-$ the parameters of a target network that stabilises learning.

Since we aim to discuss the application of this procedure towards more real-world friendly applications, we show in particular how this method helps in traffic junction environments to prevent collisions between vehicles. If agents can minimise the collisions on a more complex environment by leveraging the experience gained in a simpler one, then the agents trained are capable of making more secure decisions in these traffic scenarios. 
\begin{figure}[!t]
    \centering
    \begin{subfigure}[]{0.65\textwidth}
        \centering
        \includegraphics[width=\textwidth]{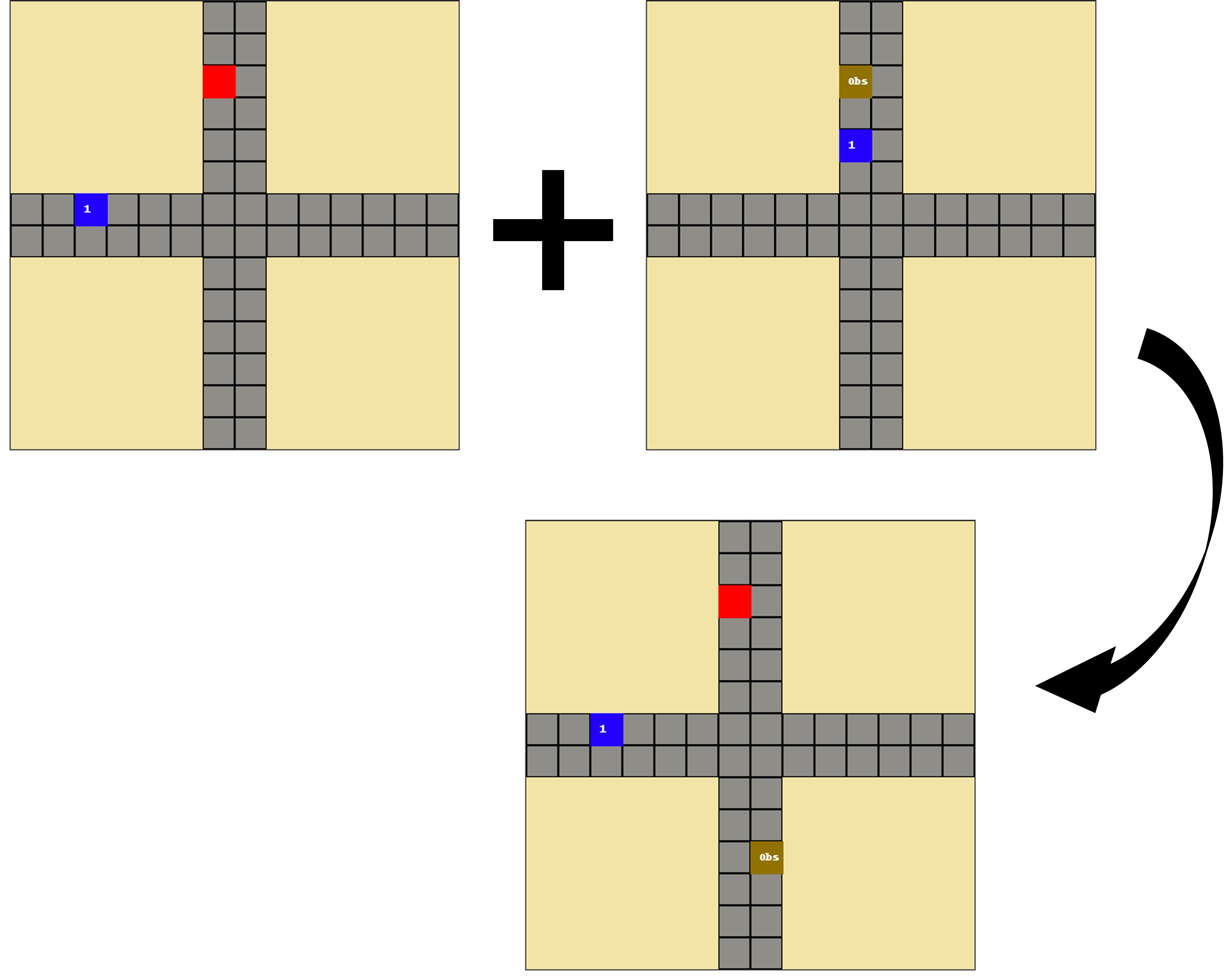}
        \caption{Environments used in section \ref{sec:exps_task_fact}: sub-tasks of goal reaching and obstacle avoidance (top, from the left to the right) and the joint task (bottom).}
        \label{}
    \end{subfigure}
    \hspace{10mm}
    \begin{subfigure}[]{0.25\textwidth}
        \centering
        \includegraphics[width=\textwidth]{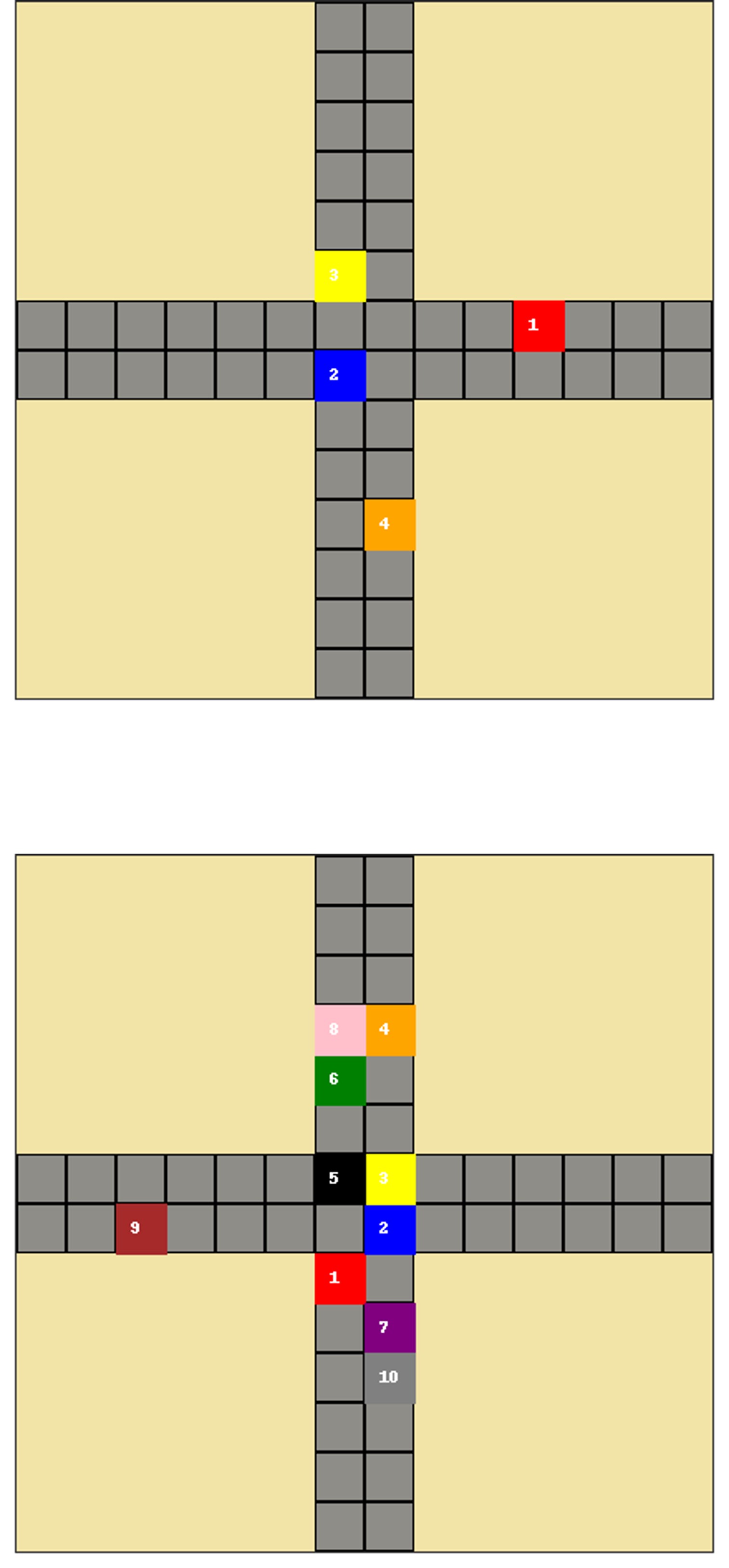}
        \caption{Environments used in section \ref{sec:exps_ctde_decent}: Traffic Junction with 4 agents (top) and 10 agents (bottom).}
        \label{}
    \end{subfigure}
    \caption{Environments used in the experiments. Although they all are based on a traffic junction, their settings are different according to the experiments in sections \ref{sec:exps_task_fact} and \ref{sec:exps_ctde_decent} (see the sections for details).}
    \label{fig:tj_maps}
\end{figure}

\section{Experiments and Results}\label{sec:exps}
In this section we conduct a set of experiments to support the hypotheses introduced in this paper. In simulated traffic junction environments we show that, in single agent settings, certain tasks can be decomposed in simpler sub-tasks that are then combined to solve a more complex task. In multi-agent settings, we investigate the discussed training structure that involves leveraging experience from CTDE and using it to help in more complex fully decentralised settings. Note that, while in both subsections of experiments we use a Traffic Junction environment, this environment has different configurations in the different experiments that we clarify ahead.

\subsection{Task Factorization}\label{sec:exps_task_fact}
\begin{figure}[!t]
    \centering
    \begin{subfigure}[]{0.40\textwidth}
    \includegraphics[width=\textwidth]{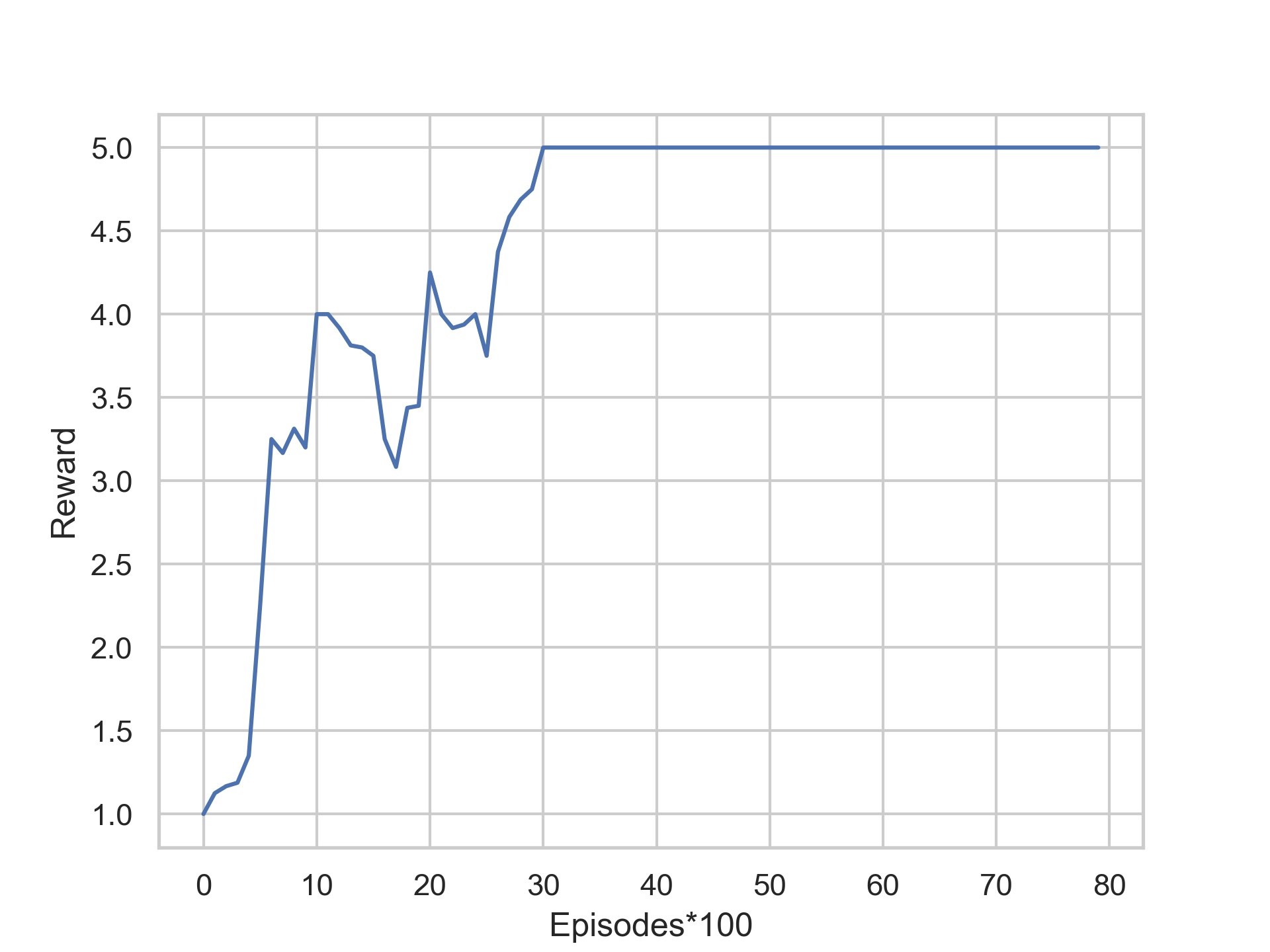}
    \caption{Sub-task 1: Goal reaching.} 
    \end{subfigure}
    \begin{subfigure}[]{0.40\textwidth}
    \includegraphics[width=\textwidth]{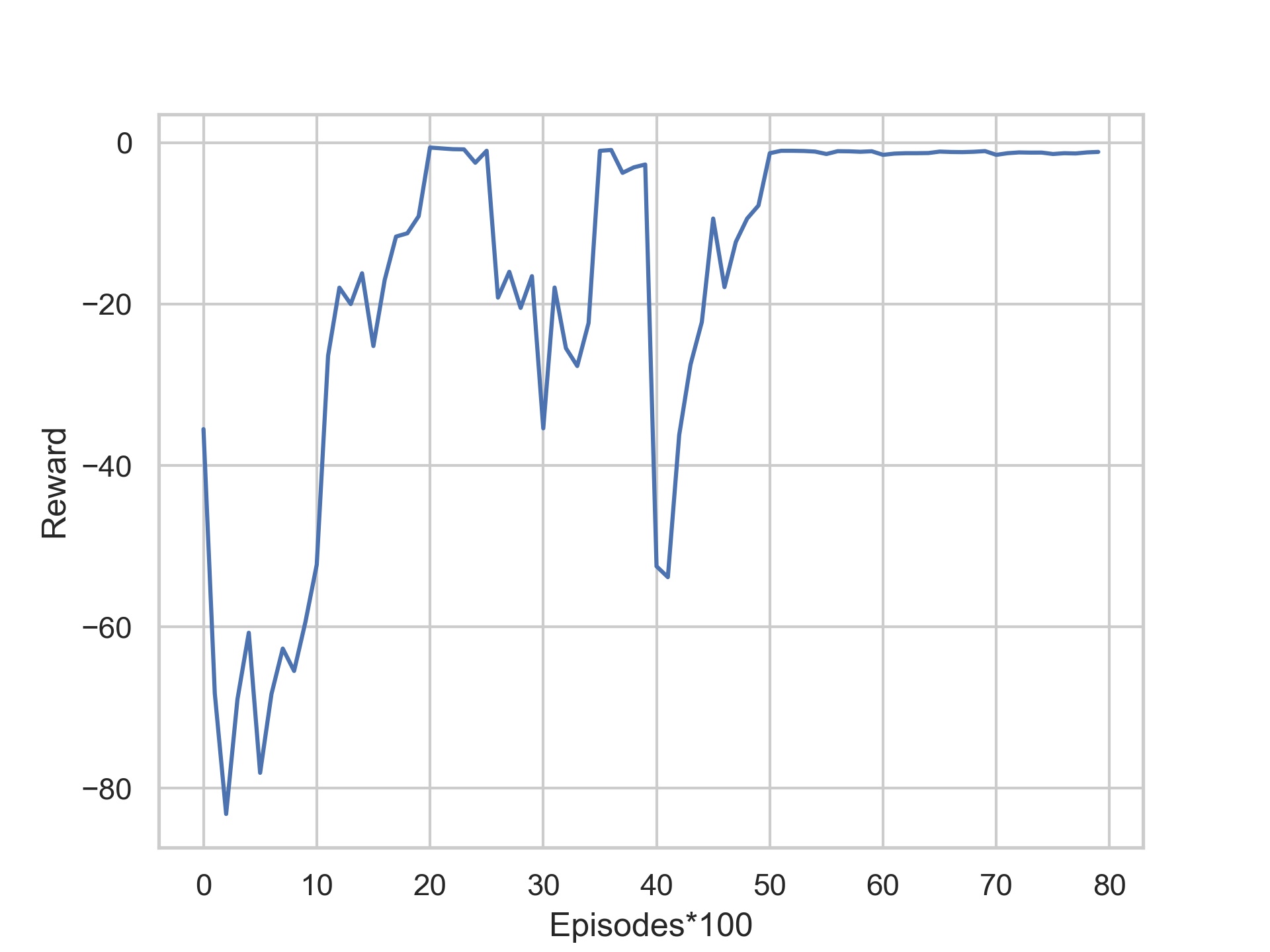}
    \caption{Sub-task 2: Obstacle avoidance.} 
    \end{subfigure}
    \begin{subfigure}[]{0.40\textwidth}
    \includegraphics[width=\textwidth]{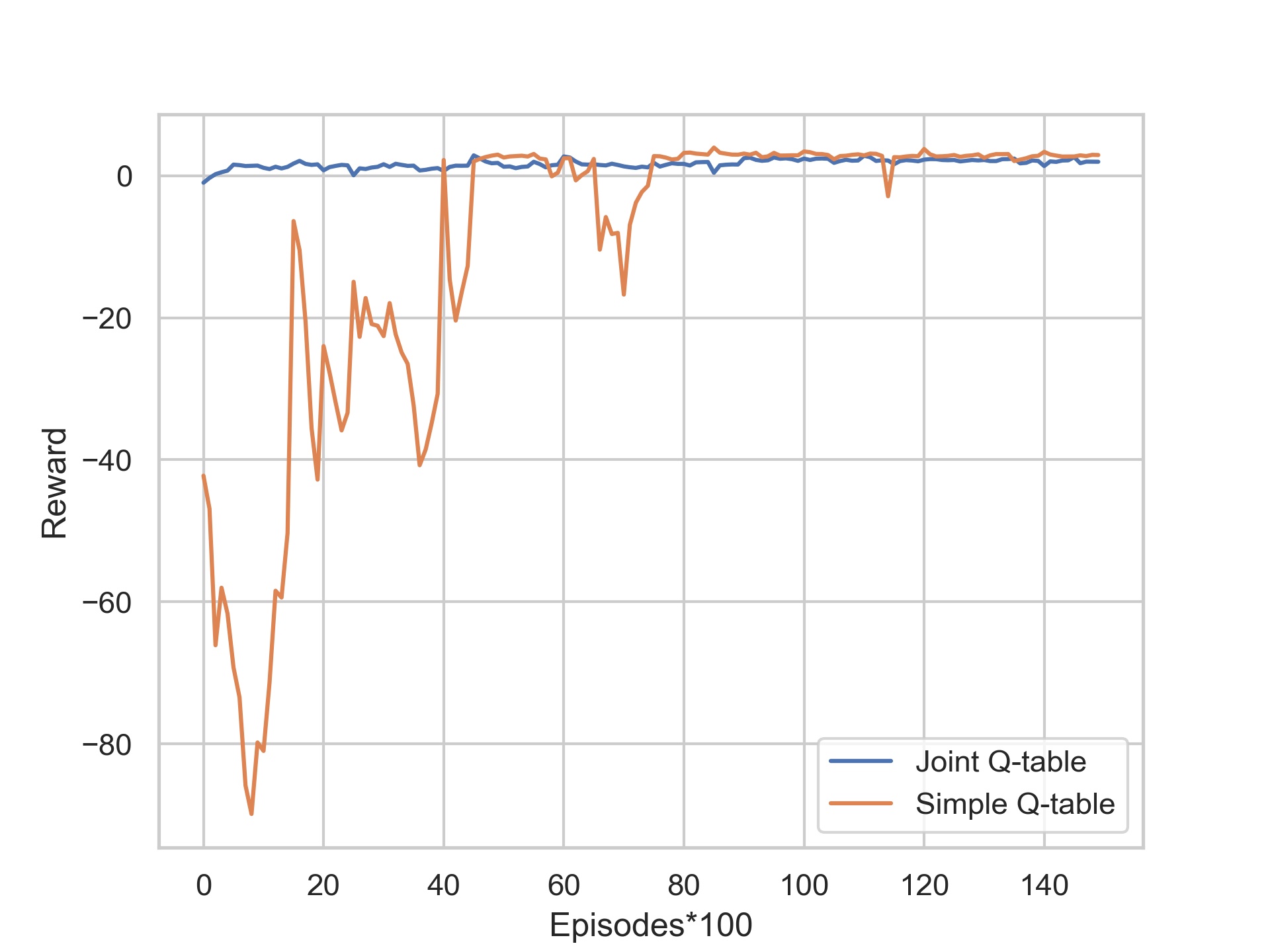}
    \caption{Main task: goal reaching with obstacle avoidance.} 
    \end{subfigure}
    \caption{Rewards achieved in the sub-tasks (upper plots) and in the complex task (bottom) as described in section \ref{sec:exps_task_fact}. The rewards are smoothed with a moving average with a 5-step sliding window.}
    \label{fig:plots_task_dec}
\end{figure}

The experiments in this subsection aim to show that a complex task based on a traffic junction (Figure \ref{fig:tj_maps}a) can be learned by breaking it in two simpler sub-tasks. In the used environment in this subsection, the task involves one agent (blue square in Fig. \ref{fig:tj_maps}a) that enters the road through one out of four possible roads and has to go to a goal place (red box in Fig. \ref{fig:tj_maps}a), while avoiding a second vehicle (brown square in Fig. \ref{fig:tj_maps}a) that is harcoded to try to collide with the agent and thus it has to avoid the collisions. The agent receives a reward of +5 when it reaches the goal and a punishment of -0.2 when it collides with the other vehicle. In addition, note that the environment is not fully observable, meaning that the agent only sees its own location and has a 3$\times$3 observation mask to observe the surroundings. In line with our needed assumption for this experiment, this task can be broken down in two sub-tasks: 1) going from the starting location to the goal (without any obstacles or other vehicles to avoid) (top left in Fig. \ref{fig:tj_maps}a) and 2) roam around the road while avoiding collisions with the hardcoded vehicle that tries to collide and must be avoided (without any goal place to find) (top right in Fig. \ref{fig:tj_maps}a). Fig. \ref{fig:plots_task_dec}a and \ref{fig:plots_task_dec}b illustrate the performance of the learned Q-tables on the 2 described sub-tasks for this scenario and Fig. \ref{fig:plots_task_dec}c shows the combination of the tables (joint Q-table) compared against a Q-table trained directly in the complex environment (we refer to this as simple Q-table). The illustration of the complex task that results from the sub-tasks can be found in the bottom of \ref{fig:tj_maps}a.

By looking at the figures, we can see that the agent solves easily the simpler tasks, achieving optimal rewards after some time. However, when looking at the more complex task that is a combination of the two simpler ones, we can see that an agent trained with the simple Q-table takes longer to learn. Although it eventually learns the task, it takes some time to get there and, if this would happen in a real scenario, the agent would cause a lot of physical damage before learning the task successfully. On the other hand, when we use the joint Q-table that mixes the two simpler learned tasks as per Algorithm \ref{alg:alg_1}, we can see that the agent can reuse what was learned in the simpler tasks and use that experience to learn much faster the complex one, avoiding the collisions that incur when using the simple Q-table directly. The results presented enhance that simple tasks can be learned, and the experience can be transferred to complex tasks. This can facilitate solving harder problems by breaking them in easier sub-tasks that can be more easily reproduced rather than stepping directly to a complex task that requires more effort and may cause problems. Also by doing so, the agent avoids many crashes that could happen by training in the complex task blindly.
\begin{figure}[!t]
\centering
\begin{subfigure}[]{0.8\textwidth}
    \centering
    \includegraphics[width=\textwidth]{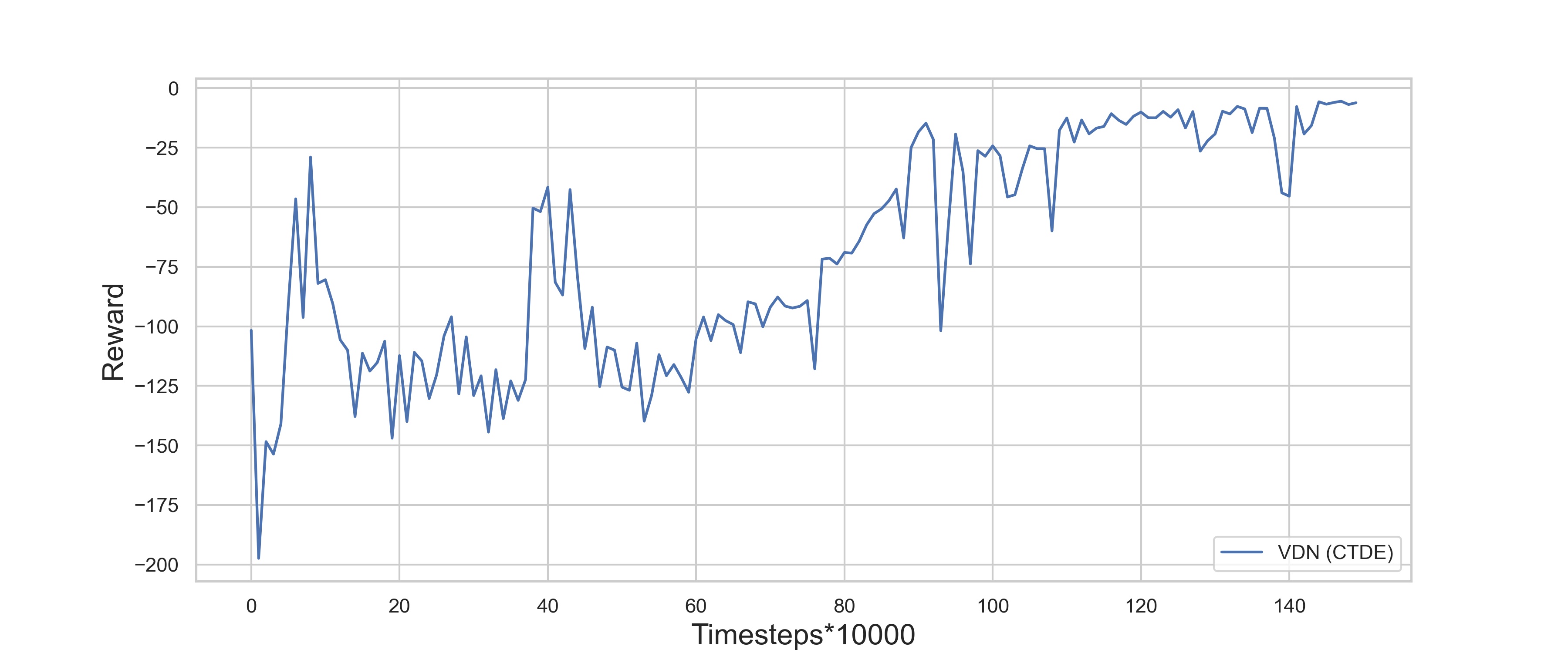}
    \caption{VDN in Traffic Junction with 4 agents.}
    \label{}
\end{subfigure}
\begin{subfigure}[]{0.8\textwidth}
    \centering
    \includegraphics[width=\textwidth]{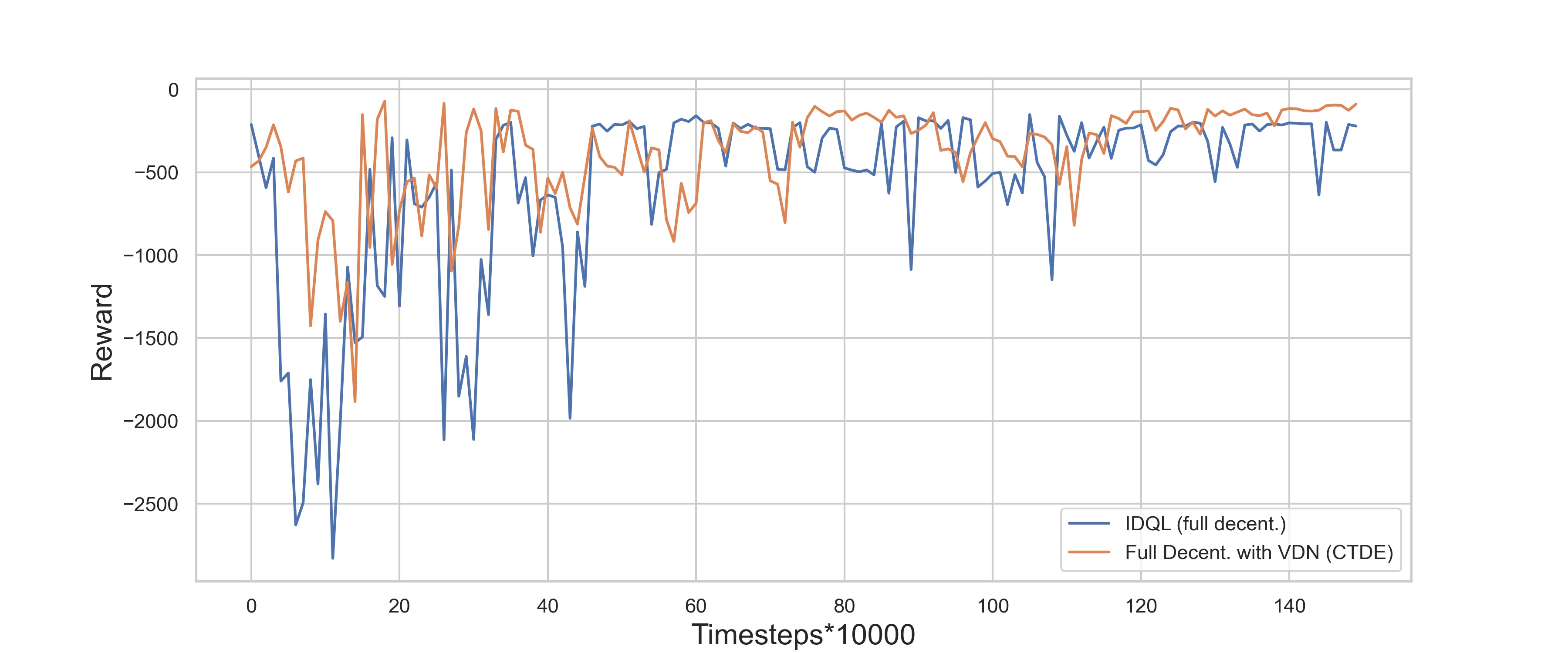}
    \caption{Traffic Junction with 10 agents for IDQL vs IDQL boosted with pre-trained VDN (in the 4 agents environment).}
    \label{}
\end{subfigure}
\caption{On the top, the rewards on the simple Traffic Junction with 4 agents (Fig. \ref{fig:tj_maps}, top), using VDN. On the bottom, the rewards for independent deep Q-learning (without parameter sharing) when starting with the policies learned in the simpler task with 4 agents using VDN (top), but now in a 10-agent scenario (Fig. \ref{fig:tj_maps}, bottom). The latter is compared against fully independent deep Q-learners (IDQL, also without parameter sharing) in the same 10-agent environment, but without the VDN starting boost.}
\label{fig:plots_ctde}
\end{figure}

\subsection{CTDE to Decentralisation}\label{sec:exps_ctde_decent}
As discussed in the previous sections, this training scheme aims to train a team of agents to learn how to negotiate their passages in a traffic junction environment (Fig. \ref{fig:tj_maps}b) while avoiding collisions with other agents. As mentioned before, note that, in this set of experiments, the configurations of the environment are different from the previous. In this set of experiments, we consider multi-agent settings where the goal is for each agent to reach a pre-defined destination at one of the other ends of the junction that is assigned to them. Each agent receives a punishment of -10 if it collides with another agent and also a penalty of -0.01 every step to incentivize them to reach the goal as fast as possible. In addition, the environment is also partially observable, and thus each agent only sees its own location, together with a 3$\times$3 observation mask to observe the surroundings, and a step counter. Following the training structure described in section \ref{sec:meth_ctde_decent}, we start by training VDN in the traffic junction environment with 4 agents. When using 4 agents, the environment is relatively simple to solve, and thus we can see that convergence is reached and remains until the end of training (Fig. \ref{fig:plots_ctde}a). By observing Fig. \ref{fig:plots_ctde}a, we can say that the method has reached an optimal reward at the end of training and the agents have learned how to solve the task. Because the method also uses the parameter sharing convention to speed up training, at the end of training we end up with only 2 different trained networks: a mixing network and a policy network that is shared by all the agents (as summarised previously in Fig. \ref{fig:param_share_vs_no_param_share} and \ref{fig:ctde_overview}). 

At the end of training, the trained policy can be transferred to a more complex environment to be used without a mixing network, i.e., in a fully decentralised fashion without using CTDE. Importantly, since we intend to increase the number of agents in the more complex environment to demonstrate how this training scheme aids in scaling to more agents, in the simpler environment we must pad the policy network to be able to accommodate the additional agents in the harder environment. Then, this network is replicated across the independent agents to boost the training. As Fig. \ref{fig:plots_ctde}a shows, we can see that using padding does not affect the performances in the 4-agent environment since they still show an increasing performance that converges during training. To make the scenario as close to reality as possible, we do not use parameter sharing in the fully decentralised version. This allows us to treat each agent as an isolated unit that does not have any access to any extra information besides their own individual observations of the environment. In Fig. \ref{fig:plots_ctde}b we can see that, when we use the transferred policy networks in the harder environment with 10 agents, the team improves the performances when compared to when they would have not taken advantage of CTDE before. In fact, we can see that, in the latter, agents take much longer to achieve a higher reward, meaning that they would cause much more collisions in the traffic environment before learning. Besides taking longer to learn, with simple fully independent learning the agents fail to achieve as good performance at the end of training as the ones with the prior boost of using policies that were trained before in the advantageous CTDE. These results show that, although CTDE is mostly only feasible in simulation, we can still take advantage of its potential to be used to improve learning in less advantageous scenarios that relate more to reality-compatible applications, helping to prepare agents for more complex situations than what they are used to, and minimising potential safety-critical problems. 

\section{Conclusion and Further Work}\label{sec:conc}
While RL has shown remarkable advances in simulated environments, approximating it to solve real problems is still challenging. In this paper, we have investigated two concepts that, if successfully applied to simulated environments, can ease the reuse of simulated RL in reality. Executing complex tasks in reality can be facilitated if these tasks are broken down to simpler sub-tasks that have been solved before. Logically, mapping simpler tasks in simulation is easier than start solving a complex task straight away. From a multi-agent perspective, with the increasing complexity of multi-agent problems, there is a need to take all the possible advantage from simulated environments. In this sense, it is important to understand how state-of-the-art methods that solve complex simulated environments can be efficiently leveraged to aid in real scenarios. With our preliminary experiments, we introduce how these can be used to tackle problems that are conceptually closer to real scenarios, although still in simulation.

In the future, we aim to study other ways of combining sub-tasks, and how tasks can also be decomposed for multi-agent settings in an automated manner. Furthermore, we intend to extend the described training structuring mechanism for MARL directly from simulation to real-world scenarios. We also intend to study how we can better preserve the learned performances from one stage to the other in this mechanism.

With this work, we aim to incentivise further research on how both single-agent and multi-agent reinforcement learning methods can be trained in simulation and then mapped to reality with reduced losses of performance. While the presented experiments are still in simulation, we believe that these preliminary results can inspire further research in this direction.

%
%
%
\bibliographystyle{splncs04}

%

\end{document}